\documentclass{article}

\usepackage[preprint]{neurips_2026}

\usepackage[utf8]{inputenc}
\usepackage[T1]{fontenc}
\usepackage{url}
\usepackage{booktabs}
\usepackage{amsmath}
\usepackage{amsfonts}
\usepackage{nicefrac}
\usepackage{microtype}
\usepackage{xcolor}
\usepackage{graphicx}
\usepackage{placeins}
\usepackage{hyperref}

\title{Measuring Annotation Efficiency for Handwritten Devanagari Recognition: Sample-Complexity Curves for Four Pretraining Regimes}

\author{
  Manglesh Kumar Pandey \\
  Alliance School of Liberal Arts and Sciences,\\ Alliance University \\
  Bengaluru, India
  \And
  Sumit Kumar Banshal \\
 Department of Computer Science and Engineering,\\ Alliance University \\
  Bengaluru, India
}
\begin{document}

\maketitle

\begin{abstract}
To train handwritten text recognition systems we need word images and their
corresponding transcriptions, and these transcriptions are produced manually.
For a script that can be read by only a small number of specialists, this
manual transcription is a limitation, because the trained models are supposed
to save the time of those same specialists. A relevant question therefore
arises: how many transcriptions are needed before a recogniser becomes useful,
and how much of that cost can pretraining remove? In this study the answer is
measured directly for handwritten Devanagari. We keep the recogniser, optimiser
and evaluation protocol the same and change only the number of real transcribed
words used for fine-tuning, across nine budgets from 10 to 4{,}000 and four
initialisation regimes, with six seeds at every point. The resulting curves are
then converted into annotation-equivalent terms. A CER of 0.50 is reached by
supervised synthetic pretraining using only 81 transcribed words, whereas
random initialisation requires 355, which gives a label multiplier of
$4.40\times$ $[3.56, 4.99]$. There is a zero-shot reference point as well: with
no real transcribed words at all, this pretraining is worth about 136 of them.
This advantage gets smaller as the target accuracy improves, and at the most
demanding target we measure, it cannot be distinguished from no saving at all.
A fourth arm in which only the encoder is transferred separates the effect of
the pretraining method from that of transfer scope, and masked image modelling
is observed to transfer negatively over a bounded range of budgets. We
emphasise that the scarcity in this study is constructed by subsampling a large
corpus.
\end{abstract}

\section{Introduction}
\label{sec:introduction}

Images of handwritten words along with their transcriptions are used to train
handwritten text recognition systems and this is done manually. For widely
written scripts this cost can be absorbed once and then used across many users
and projects. For a script that can be read by few dozen specialists, the
manual transcription is a limitation: the recognition system is supposed to save
time of these same specialists who are responsible for creating the
transcriptions.

This creates an unusual practical question. Not \emph{how accurate can a
recogniser be}, but \emph{how much transcription is needed before it becomes
useful}, and \emph{how much of that cost can be avoided}. Synthetic pretraining,
self-supervised pretraining, transfer from related scripts are some of the
methods which can help in reducing the amount of transcription required. The
size of that reduction, however, is rarely measured across a range of annotation
budgets, and it is rarely expressed in units of human effort.

In this paper, we measure it directly by varying only the number of real
transcribed words available for fine-tuning, across nine budgets from 10 to
4{,}000 and four initialisation regimes, with six seeds at every point and
keeping the recogniser, optimiser, and evaluation protocol fixed. We then use
the resulting curves to determine how many human-transcribed words each method
needs to reach a given level of accuracy. This tells us how much annotation work
can be saved by using pretraining.

Our contributions are:

\begin{itemize}
  \item A sample-complexity curve for handwritten Devanagari word recognition
    across four initialisation regimes and nine annotation budgets, with
    seed-resampled uncertainty at every point.
  \item An annotation-equivalent conversion of those curves. Supervised
    synthetic pretraining reaches CER 0.50 with 81 transcribed words where
    random initialisation requires 355, a $4.40\times$ label multiplier with a
    seed-resampled interval of $[3.56, 4.99]$.
  \item A zero-shot anchor: the synthetically pretrained model, with no real
    labels at all, achieves CER 0.7322, which the scratch curve reaches at
    approximately 136 transcribed words.
  \item A controlled comparison that separates the effect of the
    \emph{pretraining task} from \emph{how much of the network is transferred},
    using a fourth arm that carries over only the encoder from the supervised
    checkpoint. This reveals that masked image modelling produces negative
    transfer over a limited range ($n = 50$--$500$), and that the encoder and
    sequence-model parts of the supervised benefit behave differently: the
    encoder part is small but remains detectable up to $n = 1000$, while the
    much larger sequence-model part is no longer detectable beyond $n = 250$.
\end{itemize}

All the methods used in this work already exist. Our contribution is therefore
not a new method but a way of measuring the effect of pre-existing methods. We
provide a fine-grained curve in the annotation-starved regime, with
seed-resampled uncertainty intervals, expressed in units of human effort.

\section{Related Work}
\label{sec:related}
There has been a systematic study of the relationship between volume of
training data and recognition accuracy for scene text, where dense sampling is
feasible because of the existence of large synthetic and real corpora
\citep{rang2024}. That work studies behaviour far above the regime we work in,
and measures accuracy as a function of data. Scaling curves have been studied
more broadly in machine learning to describe how the availability of more data
changes performance \citep{viering2023}. None of this work expresses the
relationship in the other direction, as the number of labels required to reach
a target accuracy. Our work focuses on the annotation-starved regime and
expresses the resulting curves in terms of human transcription effort.

Using rendered text to train text recognisers is well established practice for
scene text \citep{jaderberg2014}, and synthetic data have also been used for
handwritten text recognition (HTR) in under-resourced settings, including
Devanagari \citep{dutta2018offline}. More recently, \citet{pippi2023synth} have
examined synthetic pretraining for handwriting processing tasks, and
\citet{wolf2024} synthetic-to-real adaptation via self-training. Separately,
\citet{souibgui2022oneshot} explored generating training data from very few
examples. This literature establishes that synthetic pretraining helps, and one
line of work measures how the benefit varies with the amount of real
fine-tuning data: \citet{pippi2023choose} fine-tune a CRNN on progressively
smaller fractions of three single-author historical collections and identify
which pretraining corpus best suits a given target. Their independent variable
is the choice of pretraining source, and their from-scratch baseline does not
converge at their smallest budgets, so the comparison stops at error rates
measured at fixed fractions. We instead fix the pretraining source, vary how
much of the network is transferred, retain a from-scratch baseline at every
budget, and convert the resulting curves into the number of human
transcriptions each regime saves.

How much of a pretrained network is worth carrying over is an important
question for text recognition. \citet{yosinski2014} showed that the generality
of learned features varies across model layers, and that transferability
decreases as layers become more task specific. They also attribute part of the
transfer gap to optimisation difficulties introduced by splitting a network.
Our experiment asks the same question for a CRNN trained with CTC, at each
point along an annotation budget, and expresses the answer in terms of
annotation efficiency rather than transferability alone.

Masked modelling has been applied to text recognition \citep{lyu2022maskocr}
and to self-supervised pre-training of text recognisers more broadly
\citep{kiss2024}, where it was effective but did not consistently outperform
transfer learning from closely related domains. We include masked image
modelling and supervised synthetic pretraining as arms in this study under the
same compute budget, and report the result we obtain rather than the result the
literature would predict.

Few-shot and low-resource approaches to handwritten text recognition have
received sustained attention \citep{souibgui2022few}. For Indic scripts, the
dataset we use was released with the original Devanagari benchmark
\citep{dutta2018offline} and later incorporated into a ten-script collection
\citep{gongidi2021}; competition results across Indic scripts provide
additional context \citep{mondal2023}. We emphasise that Devanagari handwriting
is not itself under-resourced: our scarcity is constructed by subsampling, for
reasons set out in Section~\ref{sec:dataset}.

\section{Experimental Setup}
\label{sec:setup}

\subsection{Dataset}
\label{sec:dataset}

A handwritten Devanagari word dataset was published in 2018
\citep{dutta2018offline}, as IIIT-HW-Dev, and in this experiment, we
specifically use IIIT-HW-Dev v1, which corrects the segmentation errors in the
initially released version. This dataset has now been incorporated into the
ten-script IIIT-INDIC-HW-WORDS collection \citep{gongidi2021}. It contains
95{,}430 word images from 12 writers, which is further divided into training
(69{,}853 images, 7 writers), validation (12{,}708 images, 2 writers), and test
(12{,}869 images, 3 writers). We verified disjointness directly from the image
paths; each split, i.e., train, test, and validation, has different writers.

All labels are NFC-normalised before use. In the released corpus, 2{,}488
labels are not in NFC form, arising from eight Devanagari nuqta letters that
Unicode admits in both precomposed and decomposed forms. CER would be measured
partly as a function of encoding artefacts, as without normalisation, the same
visual word yields two distinct label strings. To measure the geometry of the
images, we sampled a random subset of 3{,}000 images per split and measured the
following: median height is 292--293 pixels, median aspect ratio is 2.5, and
the 99th percentile of aspect ratio is 4.5--4.6. There are 100 distinct
characters in the character inventory, giving 101 CTC classes including the CTC
blank class.

We would like to emphasise that Devanagari handwriting is not a low-resource
setting, and we did not select it to represent any other script that is
low-resource. In this experiment, we manufactured scarcity by subsampling a
corpus of 95{,}430 transcribed words. The alternative is measuring
sample-complexity curves on a genuinely low-resource script, but a sufficiently
large corpus is needed to establish a full-data reference point and
systematically evaluate progressively smaller subsets. Thus, we must make it
clear that the multipliers derived as a result of this experiment apply to this
corpus, and we do not claim that they transfer.

\subsection{Recognition Model}
\label{sec:model}

A CRNN \citep{shi2017} trained with CTC loss \citep{graves2006} is common among
all four arms, i.e., \texttt{scratch}, \texttt{mim}, \texttt{synthenc}, and
\texttt{synthsup}. All input images are converted to grayscale and resized to
height 64 keeping the aspect ratio, and are then right-padded to width 320. For
the padding we use the 90th percentile of the image's own intensity and not
black, so that the padded region reads as background and not as a block of ink.
To prohibit paper tone and pen darkness from carrying the writer's identity to
input all images are standardised individually instead of by an affine
transformation, this is important because the test writers, the validation
writers and the training writers are disjoint.

The image height is reduced to 1, and width is divided by four to give 80 time
steps by the convolutional encoder. The sequence model is a two-layer
bidirectional LSTM with hidden size 256 and dropout 0.1 and then followed by a
linear CTC head. The network split is relevant because the transfer analysis in
Section~\ref{sec:transfer}, compares encoder-only and full-model transfer, the
split is as follows: total 10{,}067{,}813 parameters, of which 68.2\% belong to
the convolutional encoder, 31.3\% to the recurrent layers, and 0.5\% to the
output head.

We measured a sample of 3{,}000 images across six configurations of input
height and width stride, and checked whether the encoder produces enough
horizontal positions for CTC to align the label. The chosen configuration,
height 64 with stride 4, never ran out of positions and leaves 4.44 time steps
per character at the 1st percentile of the time-steps-per-label distribution.
The most aggressive configuration, height 32 with stride 8, leaves only 1.09 and
fails outright on 7 of the 3{,}000 images. Therefore, we chose the 64/4
configuration.

CTC decoding is greedy best-path with no language model and lexicon, the reason
for having no lexicon is that out of 6{,}907 distinct test words none are absent
from 9{,}495 distinct training words. Thus, the existence of a lexicon would
contain every test word, and lexicon-constrained decoding would reduce a
significant number of errors by simply using lookup rather than recognition.
Therefore, we report unconstrained decoding throughout and discuss
closed-vocabulary implications in Section~\ref{sec:limitations}.

\subsection{Synthetic Corpus}
\label{sec:synth}

20{,}000 rendered word images constitute synthetic pretraining corpus, with mean
label length of 6.26 characters. It is generated from the same 100-character
inventory as the real corpus. Because natural Hindi has uneven frequency
distribution of aksharas, it would be undesirable for low-label experiments to
render synthetic words from a natural text corpus, therefore, we use an
akshara-level grammar to generate synthetic word corpus, so a broader and more
uniform character coverage can be achieved. We intentionally trade off
linguistic realism in synthetic data so that rare characters/conjuncts would get
enough training examples.

We measured the appearance of the real images and used those measurements to set
the targets of the generator. On the 3{,}000-image samples described in
Section~\ref{sec:dataset}, the median ink fraction is $0.068$--$0.075$ across
the splits, the 10th to 90th percentile of ink--background separation is
$0.261$--$0.409$, the median background level is $0.859$--$0.860$, and the 1st
to 99th percentile of aspect ratio is $1.43$--$4.59$. The generator itself was
configured with target ink fraction $0.030$--$0.058$, target separation
$0.37$--$0.50$, background level $0.86$--$0.96$, aspect ratio $1.45$--$4.60$,
and slant $-9^{\circ}$ to $+9^{\circ}$. The ink fraction and separation targets
are deliberately offset from the measured values, because the generator
estimates these two quantities in a different way from the measurement described
in Section~\ref{sec:dataset} and the two estimates do not agree. Synthetic
labels require no human transcription and are therefore never counted in the
annotation budget.

\subsection{Training Regimes}
\label{sec:regimes}

We compare four arms to study the effects of pretext task and transfer scope,
along with a randomly initialised control.

\begin{table}[ht]
\centering
\caption{The four training regimes. \emph{Transferred} names the parameter
groups initialised from a pretraining checkpoint; \emph{Tensors} is the number
of state-dict entries carried over.}
\label{tab:arms}
\begin{tabular}{llll}
\toprule
Arm & Pretext task & Transferred & Tensors \\
\midrule
\texttt{scratch}  & none                     & none (random init)      & 0  \\
\texttt{mim}      & masked image modelling   & encoder                 & 29 \\
\texttt{synthenc} & supervised synthetic CTC & encoder                 & 29 \\
\texttt{synthsup} & supervised synthetic CTC & encoder + LSTM + head   & 47 \\
\bottomrule
\end{tabular}
\end{table}

The \texttt{synthenc} arm separates the two factors. Without it, any gap between
\texttt{mim} and \texttt{synthsup} would confound the type of pretraining with
the transfer scope. We retain only the convolutional parameters from the
\texttt{synthsup} checkpoint for \texttt{synthenc}, so the two supervised arms
share identical pretraining and differ only in how much of the network is
carried over.

Our masked-image-modelling arm follows the design of SimMIM
\citep{xie2022simmim}: masked regions are passed to the encoder rather than
removed from it, a lightweight decoder predicts raw pixel values, and the loss
is an $\ell_1$ term computed on the hidden pixels only. Random full-height
vertical strips of width 16 are hidden at a masking ratio of 0.5, and a
transposed-convolution decoder reconstructs them. We depart from SimMIM in two
respects. First, SimMIM uses a vision transformer with learnable mask tokens,
whereas our encoder is convolutional and masked pixels are simply set to zero.
Second, SimMIM masks square patches, whereas we mask full-height strips, because
handwriting is a horizontal sequence: a full-height strip removes roughly one
character, forcing the model to infer it from neighbouring context rather than
from local texture.

Both pretraining modes run for 20 epochs with AdamW \citep{loshchilov2019} and a
one-cycle schedule \citep{smith2017} at peak learning rate $3\times10^{-4}$. The
two modes do not use the same weight decay: it is $10^{-4}$ for the supervised
pretraining and $5\times10^{-2}$ for masked image modelling. The supervised
pretraining had also converged within this budget, whereas the masked
pretraining had not. Both of these differences bias the comparison against
\texttt{mim}. We disclose them rather than correct for them, since extending
\texttt{mim} training would break the equal-compute comparison in the opposite
direction.

\subsection{Annotation-Budget Protocol}
\label{sec:protocol}

All four arms are fine-tuned at nine label counts,
$n \in \{10, 25, 50, 100, 250, 500, 1000, 2000, 4000\}$, with six seeds at every
point, giving 36 arm-by-budget cells and 216 fine-tuning runs. We use
comparatively more label counts at the lower end as we are interested in
observing the data-scarce setting, while the larger label counts help to show
how the performance changes as more labelled data becomes available. To compare
the results fairly, the same labelled subset is used for all four arms at a
given $n$ for each seed $k$.

Optimisation uses AdamW at a learning rate of $3\times10^{-4}$, weight decay
$10^{-4}$, batch size $\min(32,n)$, gradient clipping at 5.0, and a one-cycle
schedule over a 2500-step budget. We use FP32 to calculate the CTC loss, while
most of the network uses FP16 to make training faster and use less memory; in
FP16, the log-sum-exp over the alignment lattice underflows and the loss
silently becomes NaN. Slight photometric augmentation is applied to real
training images only.

\subsection{Evaluation and Model Selection}
\label{sec:evaluation}

We evaluate the metric using corpus-level CER over a constant 4{,}000-word
sample of the test split, across all four arms and all label counts.

The 1{,}500-word validation subset is also identical across arms and is
evaluated for CER and loss every 125 steps, with early stopping if six
consecutive checks show no improvement. On the x-axis, we count only the
training labels and do not include the validation budget when reporting the
label budget. In a real low-resource setting, a validation set of such magnitude
would itself be an additional annotation cost, especially at low budgets like
$n=10$, where it exceeds the training set by two orders of magnitude.

The question therefore arises: \emph{could our unusually large validation set be
responsible for the results?} We therefore tested this directly. Using a
validation set of size $\max(n,10)$, all four arms were re-run at
$n \in \{10, 50, 100, 250, 500\}$ with three seeds, forcing the validation set
not to exceed the training set. This gave 20 arm-by-budget points in total,
reported in Table~\ref{tab:valbudget}. The largest change in mean test CER was
$+0.0125$ for \texttt{synthsup} at $n=10$, while the seed standard deviation at
this point was $0.0129$. This change is within the normal variation between
seeds, but we still prefer to report the margin rather than describe the result
as comfortable. At every other point the change was at most $+0.0075$, no point
changed by more than its seed standard deviation, and the ranking of the arms
was preserved at every budget.

\begin{table}[t]
\centering
\small
\caption{Effect of scaling the validation budget with the training budget. Side
runs use a validation set of $\max(n,10)$ words with three seeds; each is
compared seed-for-seed against the first three seeds of the main sweep at the
same $n$. The final column reports whether $|\text{diff}|$ is smaller than the
full six-seed standard deviation of the main sweep at that point. Source:
\texttt{validation\_budget\_robustness.csv}.}
\label{tab:valbudget}
\begin{tabular}{lrrrrrl}
\toprule
Arm & $n$ & Scaled val. & Fixed val. & Diff & Seed s.d. & Within s.d. \\
\midrule
\texttt{scratch}  & 10  & 0.9685 & 0.9610 & $+0.0075$ & 0.0176 & yes \\
                  & 50  & 0.8966 & 0.8953 & $+0.0013$ & 0.0124 & yes \\
                  & 100 & 0.8080 & 0.8069 & $+0.0011$ & 0.0424 & yes \\
                  & 250 & 0.6080 & 0.6064 & $+0.0016$ & 0.0323 & yes \\
                  & 500 & 0.4136 & 0.4121 & $+0.0015$ & 0.0158 & yes \\
\midrule
\texttt{synthsup} & 10  & 0.6619 & 0.6495 & $+0.0125$ & 0.0129 & yes \\
                  & 50  & 0.5626 & 0.5591 & $+0.0035$ & 0.0120 & yes \\
                  & 100 & 0.4578 & 0.4548 & $+0.0030$ & 0.0281 & yes \\
                  & 250 & 0.4446 & 0.4403 & $+0.0042$ & 0.0168 & yes \\
                  & 500 & 0.3625 & 0.3625 & $-0.0000$ & 0.0257 & yes \\
\midrule
\texttt{synthenc} & 10  & 0.9314 & 0.9283 & $+0.0030$ & 0.0179 & yes \\
                  & 50  & 0.8745 & 0.8754 & $-0.0009$ & 0.0060 & yes \\
                  & 100 & 0.7491 & 0.7528 & $-0.0037$ & 0.0573 & yes \\
                  & 250 & 0.5807 & 0.5777 & $+0.0030$ & 0.0302 & yes \\
                  & 500 & 0.3931 & 0.3908 & $+0.0024$ & 0.0181 & yes \\
\midrule
\texttt{mim}      & 10  & 0.9428 & 0.9425 & $+0.0002$ & 0.0045 & yes \\
                  & 50  & 0.9212 & 0.9199 & $+0.0012$ & 0.0132 & yes \\
                  & 100 & 0.8672 & 0.8666 & $+0.0006$ & 0.0159 & yes \\
                  & 250 & 0.6713 & 0.6713 & $+0.0000$ & 0.0315 & yes \\
                  & 500 & 0.4405 & 0.4396 & $+0.0009$ & 0.0174 & yes \\
\bottomrule
\end{tabular}
\end{table}

This validation check does not show that the experiment is robust to all sources
of noise. It only shows that reducing the validation set changed the selected
checkpoint at a small number of points, because the validation set is used to
select the best checkpoint and not to train the model. Therefore, we report the
main results using the fixed validation protocol so that model selection is
constant across all four arms.

\subsection{Uncertainty}
\label{sec:uncertainty}

Uncertainty intervals are the 2{.}5th and 97{.}5th percentiles over 2{,}000
bootstrap replicates \citep{efron1979} resampling the six random seeds. Within
each replicate the same resampled seed indices are used for every arm, so
differences and ratios between arms are matched comparisons rather than
independent estimates divided against one another. Ratios are formed inside each
replicate, so that both label counts move together and instability in a flat
curve segment propagates into the ratio instead of being averaged away before
the division.

These intervals capture variation due to the choice of random seed only. They do
not account for test-set sampling, writer variation, or interpolation error. The
test split contains three writers (Section~\ref{sec:dataset}), so writer
variation in particular cannot be estimated from a single draw; we return to
this in Section~\ref{sec:limitations}. With six seeds, the intervals should be
read as indicative rather than as precise statistical estimates.

\subsection{Annotation-Equivalent Analysis}
\label{sec:annotation-equivalent}

To express the results in terms of annotation effort, we compare the number of
real labelled words each arm needs to reach a given test CER~$c$. We use
log-log interpolation between adjacent measured points to estimate this value.
Synthetic data never enters these calculations because it does not require
human transcription effort. The annotation saving relative to the scratch
baseline is
\[
  \text{saving}(c) \;=\; 1 - \frac{N_{\text{pretrained}}(c)}{N_{\text{scratch}}(c)},
\]
and the label multiplier is
\[
  \frac{N_{\text{scratch}}(c)}{N_{\text{pretrained}}(c)},
\]
where $N_{\text{scratch}}(c)$ and $N_{\text{pretrained}}(c)$ denote the number
of real labelled words at which the scratch baseline and a pretrained arm,
respectively, reach test CER~$c$.

We apply two rules when estimating these measures. First, we interpolate
between observed points only and never extrapolate beyond the measured range;
if a target CER falls outside an arm's range, we report no estimate. Second,
we interpolate only over segments where CER decreases monotonically with
increasing~$n$, because inverting a non-monotonic function would produce
ambiguous annotation-equivalent estimates.

\FloatBarrier
\section{Annotation-Efficiency Curves}
\label{sec:curves}

\begin{figure}[t]
\centering
\includegraphics[width=\textwidth]{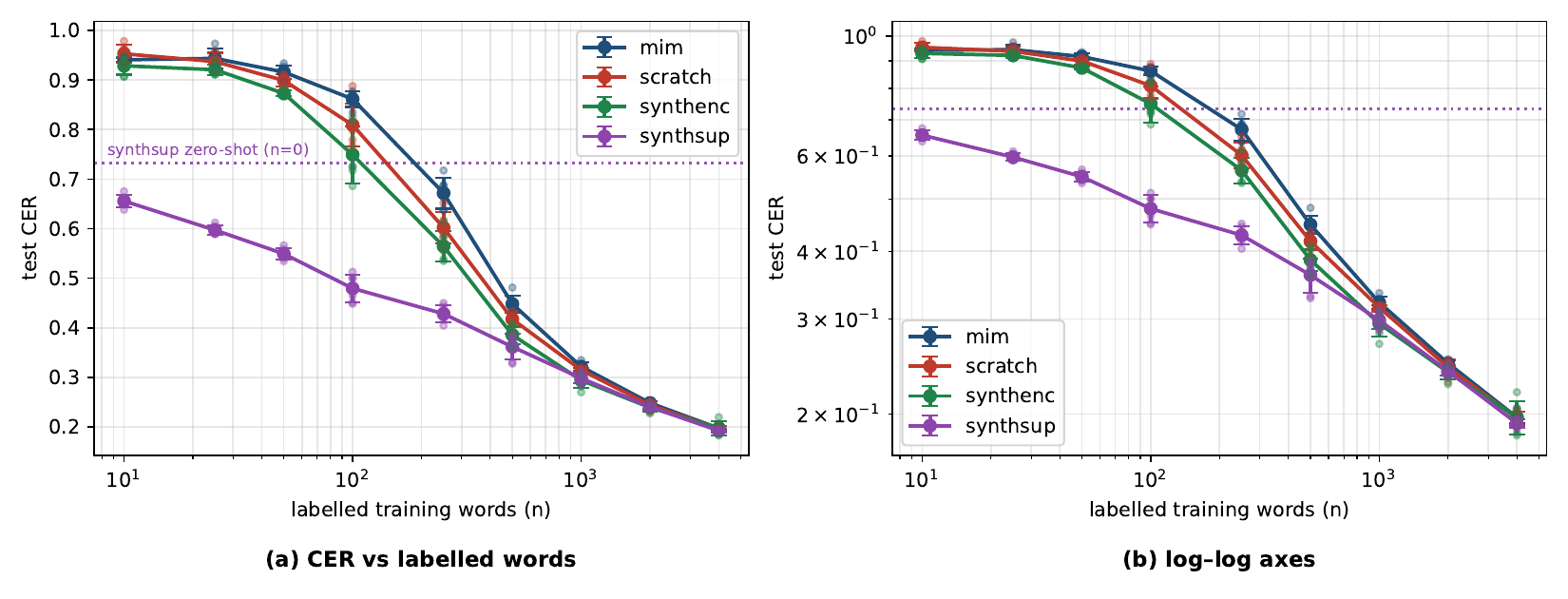}
\caption{Test CER against the number of real transcribed training words. Panel
(a) uses a logarithmic $x$-axis with a linear $y$-axis, and panel (b) uses
logarithmic axes on both sides. Each point is the mean over six seeds and the
bands show the seed standard deviation. At $n=4000$ all four arms fall between
0.1924 and 0.1975.}
\label{fig:curves}
\end{figure}

\begin{table}[t]
\centering
\caption{Mean test CER over six seeds, with seed standard deviation, for each
arm at each annotation budget. Source: \texttt{summary.csv}.}
\label{tab:summary}
\begin{tabular}{rllll}
\toprule
$n$ & \texttt{scratch} & \texttt{mim} & \texttt{synthenc} & \texttt{synthsup} \\
\midrule
10   & 0.9529\,$\pm$\,0.0176 & 0.9407\,$\pm$\,0.0045 & 0.9287\,$\pm$\,0.0179 & 0.6558\,$\pm$\,0.0129 \\
25   & 0.9372\,$\pm$\,0.0178 & 0.9436\,$\pm$\,0.0201 & 0.9207\,$\pm$\,0.0111 & 0.5971\,$\pm$\,0.0095 \\
50   & 0.8992\,$\pm$\,0.0124 & 0.9161\,$\pm$\,0.0132 & 0.8730\,$\pm$\,0.0060 & 0.5493\,$\pm$\,0.0120 \\
100  & 0.8092\,$\pm$\,0.0424 & 0.8616\,$\pm$\,0.0159 & 0.7492\,$\pm$\,0.0573 & 0.4796\,$\pm$\,0.0281 \\
250  & 0.6020\,$\pm$\,0.0323 & 0.6718\,$\pm$\,0.0315 & 0.5645\,$\pm$\,0.0302 & 0.4283\,$\pm$\,0.0168 \\
500  & 0.4174\,$\pm$\,0.0158 & 0.4480\,$\pm$\,0.0174 & 0.3855\,$\pm$\,0.0181 & 0.3615\,$\pm$\,0.0257 \\
1000 & 0.3148\,$\pm$\,0.0034 & 0.3216\,$\pm$\,0.0082 & 0.2940\,$\pm$\,0.0155 & 0.2982\,$\pm$\,0.0105 \\
2000 & 0.2442\,$\pm$\,0.0084 & 0.2483\,$\pm$\,0.0032 & 0.2391\,$\pm$\,0.0074 & 0.2402\,$\pm$\,0.0048 \\
4000 & 0.1966\,$\pm$\,0.0053 & 0.1973\,$\pm$\,0.0050 & 0.1975\,$\pm$\,0.0140 & 0.1924\,$\pm$\,0.0033 \\
\bottomrule
\end{tabular}
\end{table}

Table~\ref{tab:summary} shows the mean test CER and standard deviation over six
seeds for each experimental arm and annotation budget.
Figure~\ref{fig:curves} shows the same results as curves. The bands represent
variation over seeds. All numerical results in this section and the next are
taken from the output files generated by \texttt{analyze.py}.

\paragraph{Ordering.}
Supervised synthetic pretraining is the best arm at every budget from $n=10$ to
$n=500$, and its advantage is largest where labels are scarcest. At $n=10$, it
has a CER of 0.6558, compared to 0.9529 for random initialisation. The
encoder-only supervised arm is second over the same range. Masked image
modelling is the worst arm from $n=25$ to $n=2000$, performing worse than random
initialisation as well as the supervised arms. At $n=10$, however, masked image
modelling is slightly better than random initialisation (0.9407 versus 0.9529).

The differences between the arms become much smaller at higher annotation
budgets. The encoder-only arm performs slightly better and has a CER of 0.2940
and 0.2391 at $n=1000$ and $n=2000$ respectively, compared to the full-transfer
arm, which has a CER of 0.2982 and 0.2402 at the same budgets. At $n=4000$, all
four arms' CER values fall between 0.1924 and 0.1975, with differences
comparable to the seed standard deviations, and the ordering is no longer
meaningful. The benefits of pretraining observed in this study are therefore
concentrated at the low-annotation end of the curve. We do not extrapolate
beyond $n=4000$ in either direction, since an estimate outside the measured
range would rest on no observation at all.

\paragraph{One non-monotone point.}
The masked-pretraining curve does not improve monotonically at the smallest
budgets: its mean CER increases from 0.9407 to 0.9436 as we go from $n=10$ to
$n=25$. Learning curves that do not improve monotonically with more data are
documented across machine learning \citep{viering2023}. Since the curve is not
consistently improving in this region, the
$n=25$ point is excluded from the interpolation used to calculate
annotation-equivalent label counts. This avoids ambiguous estimates when the
curve is inverted. The point is still retained in Table~\ref{tab:summary} and
Figure~\ref{fig:curves}. This is the only point excluded from the
annotation-equivalent analysis.

\begin{table}[t]
\centering
\caption{Log-log slope between adjacent measured budgets. No arm is described
by a single exponent. Source: \texttt{segment\_slopes.csv}.}
\label{tab:slopes}
\begin{tabular}{lrrrr}
\toprule
Segment & \texttt{scratch} & \texttt{mim} & \texttt{synthenc} & \texttt{synthsup} \\
\midrule
$10\rightarrow25$     & $-0.0182$ & $+0.0033$ & $-0.0094$ & $-0.1024$ \\
$25\rightarrow50$     & $-0.0597$ & $-0.0427$ & $-0.0767$ & $-0.1203$ \\
$50\rightarrow100$    & $-0.1521$ & $-0.0885$ & $-0.2207$ & $-0.1960$ \\
$100\rightarrow250$   & $-0.3228$ & $-0.2716$ & $-0.3089$ & $-0.1233$ \\
$250\rightarrow500$   & $-0.5285$ & $-0.5844$ & $-0.5502$ & $-0.2447$ \\
$500\rightarrow1000$  & $-0.4069$ & $-0.4782$ & $-0.3910$ & $-0.2775$ \\
$1000\rightarrow2000$ & $-0.3665$ & $-0.3733$ & $-0.2985$ & $-0.3123$ \\
$2000\rightarrow4000$ & $-0.3130$ & $-0.3321$ & $-0.2758$ & $-0.3204$ \\
\bottomrule
\end{tabular}
\end{table}

\paragraph{No curve is a single power law.}
Table~\ref{tab:slopes} gives the log-log slope between each pair of adjacent
measured budgets. We could not derive a single exponent that describes any arm
well on its own, as the slopes change significantly across the annotation
range. For the scratch baseline, the slope ranges from $-0.0182$ on the
$10\rightarrow25$ segment to $-0.5285$ on the $250\rightarrow500$ segment,
before becoming less steep again at higher budgets. If we were to use a single
fitted exponent, it would hide this large change in the local slope. We
therefore report the slope for each adjacent segment rather than fitting one
exponent to the entire curve.

The supervised arms and the scratch baseline also have different curve shapes,
not just different CER levels. The scratch curve is nearly flat below $n=50$,
becomes steepest on the $250\rightarrow500$ segment, and then becomes less
steep again. The full-transfer arm, on the other hand, becomes increasingly
steep across most of the range, from $-0.1024$ on $10\rightarrow25$ to
$-0.3204$ on $2000\rightarrow4000$, with its steepest segment at the end; hence
we can say that the curves have different shapes, not just different
performance levels. Therefore, pretraining does not simply shift the scratch
curve to the left while preserving the same shape.

\paragraph{Robustness to the validation budget.}
To check whether the fixed validation budget was responsible for the strong
low-data results, we ran a smaller experiment, described in
Section~\ref{sec:evaluation} and reported in Table~\ref{tab:valbudget}. We found
no detectable effect, although this small robustness experiment cannot
completely rule one out. All 20 comparisons fell within the seed standard
deviation of their corresponding main-sweep values. The largest difference was a
$+0.0125$ CER shift for \texttt{synthsup} at $n=10$, compared with a seed
standard deviation of $0.0129$ for that setting.

\FloatBarrier
\section{What Pretraining Buys}
\label{sec:buys}

\begin{table}[t]
\centering
\small
\setlength{\tabcolsep}{4pt}
\caption{Annotation saving and label multiplier at each target CER, relative to
the scratch baseline. $N$ is the number of real transcribed words at which an
arm reaches the target, estimated by log-log interpolation between measured
points. Brackets give the 2.5th and 97.5th percentiles over 2{,}000
seed-resampled replicates. Saving and multiplier are two views of the same
quantity, related by $\text{mult} = 1/(1-\text{saving})$. A dash means the
target lies outside the arm's measured range and no estimate is reported.
Source: \texttt{annotation\_saving.csv}.}
\label{tab:saving}
\begin{tabular}{lrrrll}
\toprule
Arm & Target CER & $N_{\text{scratch}}$ & $N_{\text{arm}}$ & Saving & Multiplier \\
\midrule
\texttt{synthsup} & 0.80 & ---  & ---  & ---                             & --- \\
                  & 0.60 & 252  & 24   & $0.9052$ [$0.8905$, $0.9159$]   & $10.55\times$ [9.14, 11.89] \\
                  & 0.50 & 355  & 81   & $0.7725$ [$0.7189$, $0.7994$]   & $4.40\times$ [3.56, 4.99] \\
                  & 0.40 & 555  & 331  & $0.4043$ [$0.3042$, $0.4767$]   & $1.68\times$ [1.44, 1.91] \\
                  & 0.30 & 1140 & 979  & $0.1413$ [$0.0782$, $0.2072$]   & $1.16\times$ [1.08, 1.26] \\
                  & 0.25 & 1875 & 1760 & $0.0616$ [$0.0141$, $0.1139$]   & $1.07\times$ [1.01, 1.13] \\
                  & 0.20 & 3784 & 3542 & $0.0637$ [$-0.0148$, $0.1246$]  & $1.07\times$ [0.99, 1.14] \\
\midrule
\texttt{synthenc} & 0.80 & 104  & 74   & $0.2831$ [$0.1574$, $0.3315$]   & $1.39\times$ [1.19, 1.50] \\
                  & 0.60 & 252  & 205  & $0.1844$ [$0.0681$, $0.2762$]   & $1.23\times$ [1.07, 1.38] \\
                  & 0.50 & 355  & 312  & $0.1226$ [$0.0599$, $0.1849$]   & $1.14\times$ [1.06, 1.23] \\
                  & 0.40 & 555  & 468  & $0.1575$ [$0.0480$, $0.2294$]   & $1.19\times$ [1.05, 1.30] \\
                  & 0.30 & 1140 & 950  & $0.1672$ [$0.0730$, $0.2470$]   & $1.20\times$ [1.08, 1.33] \\
                  & 0.25 & 1875 & 1721 & $0.0821$ [$0.0070$, $0.1608$]   & $1.09\times$ [1.01, 1.19] \\
                  & 0.20 & 3784 & 3818 & $-0.0091$ [$-0.0760$, $0.1377$] & $0.99\times$ [0.93, 1.16] \\
\midrule
\texttt{mim}      & 0.80 & 104  & 131  & $-0.2683$ [$-0.4571$, $-0.1562$] & $0.79\times$ [0.69, 0.86] \\
                  & 0.60 & 252  & 303  & $-0.2056$ [$-0.2865$, $-0.1640$] & $0.83\times$ [0.78, 0.86] \\
                  & 0.50 & 355  & 414  & $-0.1665$ [$-0.2326$, $-0.1171$] & $0.86\times$ [0.81, 0.90] \\
                  & 0.40 & 555  & 634  & $-0.1419$ [$-0.2302$, $-0.0744$] & $0.88\times$ [0.81, 0.93] \\
                  & 0.30 & 1140 & 1205 & $-0.0567$ [$-0.1114$, $-0.0058$] & $0.95\times$ [0.90, 0.99] \\
                  & 0.25 & 1875 & 1964 & $-0.0472$ [$-0.0997$, $0.0058$]  & $0.95\times$ [0.91, 1.01] \\
                  & 0.20 & 3784 & 3837 & $-0.0140$ [$-0.0939$, $0.0671$]  & $0.99\times$ [0.91, 1.07] \\
\bottomrule
\end{tabular}
\end{table}

\paragraph{Headline saving.}
The annotation saving and label multiplier for each arm at target CER values
within its measured range are reported in Table~\ref{tab:saving}. At a target
CER of 0.50, 355 and 81 transcribed words are required by the scratch baseline
and \texttt{synthsup} respectively. This means an annotation saving of
$0.7725$, with a seed-resampled interval of $[0.7189, 0.7994]$, or a label
multiplier of $4.40\times$ $[3.56, 4.99]$. Simply put, we can say that
supervised synthetic pretraining reduces the transcription effort needed to
reach this accuracy by roughly three quarters. The reason behind reporting the
CER 0.50 row as our headline result, even though Table~\ref{tab:saving}
contains a larger multiplier of $10.55\times$ $[9.14, 11.89]$ at CER 0.60, is
that it comes from a longer and better-sampled part of both curves. At CER
0.60, \texttt{synthsup} interpolates between the $n=10$ and $n=25$ points,
which is the shortest and least densely sampled segment in the sweep. We
therefore prefer the CER 0.50 estimate as the main result and report the larger
CER 0.60 estimate as a secondary result.

\paragraph{The saving gets smaller as the target accuracy improves, and then
stops changing.}
The label multiplier for \texttt{synthsup} decreases as the CER decreases: from
$4.40\times$ at CER 0.50 to $1.68\times$ at CER 0.40 and $1.16\times$ at CER
0.30. After that the decline stops rather than continuing, as the saving is
$0.0616$ at CER 0.25 and $0.0637$ at CER 0.20. At CER 0.20 the saving interval
is $[-0.0148, 0.1246]$, which includes zero, so we cannot distinguish a real
annotation saving from no saving. The encoder-only arm is also not monotone
across the targets, as Table~\ref{tab:saving} shows. Within the range we
measure, the supervised synthetic arm and the encoder-only arm do not improve
the accuracy the model eventually reaches; instead, they allow the model to
reach a given accuracy using fewer real labels. The only difference between
these arms is that the latter provides a smaller benefit. Masked pretraining
shows the opposite pattern, requiring more labels than random initialisation
rather than fewer; its multiplier is below one at every target, and its interval
excludes one down to CER 0.30. We return to this result in
Section~\ref{sec:transfer}.

\begin{table}[t]
\centering
\caption{Exact-match analysis of the zero-shot \texttt{synthsup} model on the
4{,}000-word test sample, against the base rate of overlap between the test
sample and the synthetic pretraining corpus. Counts are over test instances; at
the level of distinct words the base rate is unchanged, at 34 of 3{,}356.
Source: \texttt{error\_analysis.json} and \texttt{synth\_overlap.json}.}
\label{tab:exactmatch}
\begin{tabular}{lr}
\toprule
Quantity & Count \\
\midrule
Test words evaluated                        & 4000 \\
\quad of which present in synthetic corpus  & 40 \\
Transcribed exactly correctly (zero-shot)   & 57 \\
\quad of which present in synthetic corpus  & 6 \\
\bottomrule
\end{tabular}
\end{table}

\paragraph{Zero-shot anchor.}
We also observed a zero-shot reference point provided by the synthetically
pretrained model. Without any real labelled words used for fine-tuning, it
achieves a test CER of 0.7322. The scratch curve reaches the same CER at
approximately 136 transcribed words. Thus, pretraining on only rendered text is
worth about 136 human-transcribed words before the model has even seen a single
real label.

We then checked whether this zero-shot accuracy is explained by memorisation of
the rendered word forms. It is not free of memorisation: the exact matches are
in fact enriched in words that appear in the synthetic corpus. Of the 4{,}000
test words evaluated, 40 also occur in the synthetic corpus, which is a base
rate of 1.0\%; of the 57 words transcribed exactly correctly, 6 do
(Table~\ref{tab:exactmatch}). Two caveats apply here. The counts are small, and
word length is a confounder, because short words are both easier to transcribe
and also more likely to be produced by the synthetic grammar. The point which
survives is one of magnitude. At most 6 of the 4{,}000 test words could have
been produced by memorisation, so the corpus CER of 0.7322, and the anchor of
136 transcribed words which is derived from it, are essentially unaffected.

\FloatBarrier
\section{Transfer Analysis}
\label{sec:transfer}

Section~\ref{sec:buys} reports what supervised synthetic pretraining is worth in
transcribed words. This section asks where that benefit comes from. The
experimental arms can be compared along two axes: which pretraining task was
used, and how much of the network was carried over from the pretraining
checkpoint (Table~\ref{tab:arms}). Comparing arms that differ along one axis at
a time separates the two effects.

\subsection{Full-Model vs. Encoder-Only Transfer}
\label{sec:transfer-scope}

\begin{table}[t]
\centering
\small
\setlength{\tabcolsep}{4pt}
\caption{Paired CER difference against the scratch baseline at the same label
count. A negative value means the arm is better than scratch. Within each
bootstrap replicate the same resampled seed indices are used for both arms, so
these are matched comparisons. Intervals that exclude zero are marked
$^{\dagger}$. Source: \texttt{cer\_difference.csv}.}
\label{tab:diff}
\begin{tabular}{rlll}
\toprule
$n$ & \texttt{synthsup} & \texttt{synthenc} & \texttt{mim} \\
\midrule
10   & $-0.2971^{\dagger}$ [$-0.3115$, $-0.2846$] & $-0.0242^{\dagger}$ [$-0.0419$, $-0.0057$] & $-0.0122^{\dagger}$ [$-0.0236$, $-0.0010$] \\
25   & $-0.3401^{\dagger}$ [$-0.3497$, $-0.3299$] & $-0.0165^{\dagger}$ [$-0.0328$, $-0.0020$] & $+0.0064$ [$-0.0066$, $+0.0214$] \\
50   & $-0.3498^{\dagger}$ [$-0.3624$, $-0.3361$] & $-0.0261^{\dagger}$ [$-0.0369$, $-0.0142$] & $+0.0169^{\dagger}$ [$+0.0028$, $+0.0302$] \\
100  & $-0.3297^{\dagger}$ [$-0.3650$, $-0.2893$] & $-0.0600^{\dagger}$ [$-0.0840$, $-0.0319$] & $+0.0524^{\dagger}$ [$+0.0224$, $+0.0764$] \\
250  & $-0.1737^{\dagger}$ [$-0.1960$, $-0.1567$] & $-0.0375^{\dagger}$ [$-0.0688$, $-0.0101$] & $+0.0698^{\dagger}$ [$+0.0609$, $+0.0769$] \\
500  & $-0.0559^{\dagger}$ [$-0.0832$, $-0.0314$] & $-0.0318^{\dagger}$ [$-0.0525$, $-0.0076$] & $+0.0307^{\dagger}$ [$+0.0163$, $+0.0496$] \\
1000 & $-0.0165^{\dagger}$ [$-0.0231$, $-0.0098$] & $-0.0208^{\dagger}$ [$-0.0343$, $-0.0084$] & $+0.0068$ [$-0.0005$, $+0.0150$] \\
2000 & $-0.0040$ [$-0.0090$, $+0.0009$] & $-0.0051$ [$-0.0137$, $+0.0023$] & $+0.0041$ [$-0.0006$, $+0.0094$] \\
4000 & $-0.0042$ [$-0.0097$, $+0.0016$] & $+0.0009$ [$-0.0096$, $+0.0122$] & $+0.0007$ [$-0.0062$, $+0.0076$] \\
\bottomrule
\end{tabular}
\end{table}

\begin{table}[t]
\centering
\caption{Sequence-model contribution: \texttt{synthenc} CER minus
\texttt{synthsup} CER at the same label count. A positive value means that
transferring the recurrent layers and the output head, in addition to the
encoder, gives a further improvement. Intervals that exclude zero are marked
$^{\dagger}$. Source: \texttt{sequence\_benefit.csv}.}
\label{tab:seqbenefit}
\begin{tabular}{rll}
\toprule
$n$ & Difference & 95\% interval \\
\midrule
10   & $+0.2729^{\dagger}$ & [$+0.2547$, $+0.2912$] \\
25   & $+0.3236^{\dagger}$ & [$+0.3137$, $+0.3361$] \\
50   & $+0.3237^{\dagger}$ & [$+0.3169$, $+0.3315$] \\
100  & $+0.2696^{\dagger}$ & [$+0.2201$, $+0.3189$] \\
250  & $+0.1362^{\dagger}$ & [$+0.1110$, $+0.1564$] \\
500  & $+0.0240$           & [$-0.0016$, $+0.0486$] \\
1000 & $-0.0042$           & [$-0.0213$, $+0.0095$] \\
2000 & $-0.0011$           & [$-0.0061$, $+0.0028$] \\
4000 & $+0.0051$           & [$-0.0048$, $+0.0147$] \\
\bottomrule
\end{tabular}
\end{table}

The only difference between the \texttt{synthenc} and \texttt{synthsup} arms is
the number of tensors transferred: 29 and 47 respectively, where
\texttt{synthenc} carries over the convolutional encoder alone and
\texttt{synthsup} additionally carries over the recurrent layers and the output
head. Both arms are initialised from the same pretraining checkpoint. The two
arms therefore use the same pretraining and differ only in how much of the
pretrained network is transferred. This allows us to study the effect of
transfer scope separately from the effect of the pretraining task, which is the
question \citet{yosinski2014} studied layer by layer for convolutional
classifiers.

Table~\ref{tab:diff} compares each pretrained arm with the scratch baseline at
the same label count. Encoder-only transfer helps on its own: \texttt{synthenc}
is better than scratch at every budget from $n=10$ to $n=1000$, with intervals
that exclude zero at all seven of those budgets. However, the improvement is
small, ranging from $0.0165$ to $0.0600$ CER. Full-model transfer is better than
scratch over the same range, but by a far larger margin, reaching a difference
of $0.3498$ CER at $n=50$. Thus, transferring the encoder alone provides a real
benefit, but it accounts for only a small part of the much larger improvement
obtained when the full pretrained model is transferred.

Table~\ref{tab:seqbenefit} compares \texttt{synthsup} and \texttt{synthenc}
directly. From $n=10$ to $n=250$, full-model transfer is better than
encoder-only transfer, with intervals that exclude zero. The difference is
largest at $n=50$, where it is $0.3237$ CER. At $n=500$, the interval is
$[-0.0016, +0.0486]$, which includes zero, and it also includes zero at every
larger budget. At $n=1000$ and $n=2000$, the point estimate is slightly
negative, meaning \texttt{synthenc} is nominally ahead of \texttt{synthsup}.
However, because the corresponding intervals include zero, we cannot conclude
that encoder-only transfer is actually better at these budgets.

Overall, at small budgets most of the benefit from supervised synthetic
pretraining comes from the components beyond the encoder, that is, from the
transferred recurrent layers and output head. The advantage of transferring
these additional components is clear up to $n=250$. At $n=500$ and higher,
however, the results do not provide clear evidence that full-model transfer is
better than encoder-only transfer.

\subsection{Masked Pretraining and Negative Transfer}
\label{sec:mim}

The \texttt{mim} and \texttt{synthenc} arms both transfer the same 29 encoder
tensors under the same compute budget, but they use different pretraining tasks.
Here transfer scope is held constant, so we can study the difference that the
choice of pretraining task creates.

The two arms behave in opposite ways. As shown in Table~\ref{tab:diff},
\texttt{synthenc} is better than scratch up to budget $n=1000$, whereas
\texttt{mim} is worse than scratch at $n=50$, $100$, $250$, and $500$, with
intervals excluding zero at all four budgets. The largest difference is
$+0.0698$ CER at $n=250$. At $n=1000$ and above, the differences are small and
the intervals include zero. The interval at $n=25$ also includes zero, and at
$n=10$ the \texttt{mim} arm is better than scratch by $0.0122$ CER, with an
interval that excludes zero. We therefore do not treat negative transfer as a
general property of masked pretraining. In our experiment, it is observed mainly
from $n=50$ to $n=500$.

In terms of annotation budget, masked pretraining does not reduce the amount of
transcription needed. At a target CER of 0.50, \texttt{mim} requires 414
transcribed words compared with 355 for scratch, which gives a negative
annotation saving of $-0.1665$ and a label multiplier of $0.86\times$
$[0.81, 0.90]$. The multiplier is below one at every target we report, but its
interval crosses one at CER 0.25 and CER 0.20, so at the two most demanding
targets we cannot establish that masked pretraining costs labels at all.

There are two important limitations to this comparison. Under 20 epochs of
training with the same compute budget, the supervised pretraining had converged,
whereas the masked pretraining had not, and the two modes also used different
weight decay (Section~\ref{sec:regimes}). Both of these may disadvantage
\texttt{mim}. We therefore interpret the result as evidence about masked image
modelling specific to our experimental design, rather than as a general
statement that masked pretraining is ineffective.

\subsection{Transfer Components and Curve Shape}
\label{sec:transfer-components}

The comparisons in the above subsections divide the benefit of supervised
synthetic pretraining into two parts. The encoder contribution stays within a
narrow band, between $0.0165$ and $0.0600$ CER from $n=10$ to $n=500$; it is
largest at $n=100$ and is no smaller at $n=500$ than at $n=25$. The
sequence-model contribution is roughly an order of magnitude larger at the
smallest budgets, rising from $0.2729$ CER at $n=10$ to $0.3237$ at $n=50$
before falling to $0.1362$ at $n=250$, and its interval already includes zero at
$n=500$ (Tables~\ref{tab:diff} and~\ref{tab:seqbenefit}). The two parts behave
quite differently.

The sequence model contributes more to the total benefit at small budgets and is
no longer detectable once $n$ reaches $500$, while the smaller encoder
contribution remains detectable up to $n=1000$ in absolute CER. We claim each
contribution only over the range where its interval excludes zero: up to
$n=1000$ for the encoder, and up to $n=250$ for the sequence model.

The two components also line up with the curve shapes reported in
Section~\ref{sec:curves}. The sequence-model contribution disappears between
$n=250$ and $n=500$, and the scratch curve reaches its steepest segment on
exactly that interval, with a log-log slope of $-0.5285$
(Table~\ref{tab:slopes}). The baseline closes most of the gap over the same
range in which the transferred sequence model stops helping.

Instead of using the percentage of the total benefit to report the
decomposition, we report it in terms of absolute CER. The encoder \emph{share},
as reported in \texttt{benefit\_decomposition.csv}, is as follows: it rises from
$8.2\%$ $[2.0\%, 13.7\%]$ at $n = 10$ to $57.0\%$ $[22.3\%, 103.2\%]$ at
$n = 500$. Now, these percentages may appear to show that the encoder
contributes progressively more, but this is not the case. The percentage rises
because the denominator, which is the total gap between \texttt{synthsup} and
\texttt{scratch}, collapses over this range, not because the encoder
contribution grows. This instability is also visible in the interval itself,
which crosses $100\%$ at $n = 500$. Thus, we do not use the percentage share of
the total benefit at this budget and henceforth. In contrast, the absolute
differences remain interpretable across the whole range.

\FloatBarrier
\section{Discussion}
\label{sec:discussion}

We have used pre-existing methods in this study, hence our contribution is not
a new method but rather how we evaluate these standard methods. Instead of
comparing these methods at a single fixed label budget, we measure their
performance across a range of annotation budgets.

\paragraph{A single number cannot summarise a pretraining benefit.}
For supervised synthetic pretraining, the label multiplier falls from
$10.55\times$ at CER 0.60 to $1.07\times$ at CER 0.20, where it can no longer be
told apart from no benefit at all; on our headline row it is $4.40\times$ at CER
0.50. Hence, a single reported improvement at one label budget describes only
that one point on the curve and not the general behaviour of the method. This is
why we report the full curves, and explain in Section~\ref{sec:buys} why CER
0.50 is used as our headline result.

\paragraph{Curve shape carries information that levels do not.}
The shapes of the curves of the four arms change as the annotation budget
increases, which means they are not simply parallel curves at different CER
levels. The \texttt{scratch} baseline does not improve significantly at the
smaller budgets and then improves rapidly in the middle of the range. In
contrast, the full-transfer arm improves more steadily and is steepest at the
highest budgets. We therefore do not describe any arm by a single fitted power
law, as this would ignore important changes in the curve shape.

\paragraph{Intervals decide the scope of every claim.}
In this study we claim only those effects whose seed-resampled interval
excludes zero. As we have discussed, the encoder contribution is smaller than
the sequence-model contribution, but it remains detectable further along the
label axis. The same logic is applied to the negative transfer from masked
image modelling, which is observed only over a certain range of label budgets.
If we were to ignore the seed-resampled intervals, these effects could easily
be reported as more general in nature than the evidence supports.

\paragraph{Shares are not a safe way to decompose a benefit.}
Presenting a component as a fraction of the total is tempting but proves
misleading in our study. The encoder's share of the total benefit rises across
the range, which at first might suggest that the encoder becomes more important
as the number of labels increases. The reason for this rise is that the
denominator, the total benefit, shrinks over the same range, so the share grows
even though the encoder's absolute contribution does not. We therefore report
both components in absolute CER and restrict share-based statements to the
range where the interval stays within $[0,1]$.

\FloatBarrier
\section{Limitations}
\label{sec:limitations}

\begin{enumerate}

\item Devanagari handwriting is not a low-resource setting by default. We
subsampled a corpus of 95{,}430 transcribed words to artificially create
annotation-starved conditions. We used such a large corpus because it is needed
to establish the full-data reference points. The multipliers that are derived
from this study apply to this corpus under subsampling. We do not claim they
transfer to a real under-resourced script, and we do not offer Devanagari as a
proxy for one.

\item All label counts reported in this study count training words only. Every
arm additionally uses a fixed validation set of 1{,}500 transcribed words for
model selection, which a real annotation-starved project would also have to
transcribe. At the smallest budgets this validation set is substantially larger
than the training set itself, so the annotation savings we report understate the
total transcription cost. Table~\ref{tab:valbudget} reports how the curves
respond when the validation budget is scaled with the training budget.

\item Listed uncertainty intervals resample six random seeds and do not account
for test-set sampling, writer variation or interpolation error. The true
uncertainty could be larger than the interval we derive, by an amount we have
not estimated in this study. As only six seeds are used, the intervals thus are
indicative in nature rather than being precise statistical estimates. Small
effects that our intervals cannot separate from zero may nevertheless exist, and
conversely an interval that excludes zero may do so by chance.

\item No arm ran at the exact label counts mentioned in Table~\ref{tab:saving}.
Log-log interpolation is used to interpolate between two measured points for
deriving estimates, therefore it rests on the assumption that there is a linear
relation between those two points at log-log scale. The reliability of the
estimate also depends on how closely the measured points are spaced.

\item Both pretraining methods were trained for 20 epochs under a common compute
budget. The supervised synthetic arm converged under this budget but masked
image modelling did not. The two pretraining modes also used different weight
decay, $10^{-4}$ and $5\times10^{-2}$, so the comparison between the two pretext
tasks is not controlled for optimisation hyperparameters. Both of these bias the
comparison against masked image modelling. Therefore, the negative transfer of
masked image modelling applies to this compute budget and this corpus only, and
only over the range of label counts where we measure it; it is not a general
statement about the effectiveness of masked image modelling.

\item The results are derived from a single CRNN with CTC on the same Devanagari
word dataset. We do not know whether the curve shape, the decomposition or the
negative transfer would survive the change of architecture, corpus or script.

\item Every distinct word in the test split also occurs in the training split:
0 of the 6{,}907 distinct test words are absent from the 9{,}495 distinct
training words. The error rates we report therefore do not measure
generalisation to unseen words, and would likely be worse on a test set
containing them. All arms share this split, so the comparison between arms is
unaffected.

\item An akshara-level grammar is used to render words for synthetic data. It
was chosen for character coverage and not linguistic realism. We therefore do
not vary the synthetic corpus or compare it with other forms of synthetic
pretraining.

\item We compare synthetic pretraining with random initialisation, but do not
include a pretraining arm using real handwritten data. We therefore cannot
determine whether synthetic data provide a better pretraining source than real
handwritten data.

\item The sequence-model contribution is clearly supported only up to $n = 250$,
while the encoder contribution remains supported up to $n = 1000$. We therefore
restrict claims about these two components to those ranges. Percentage-share
interpretations are restricted further to cases where the uncertainty interval
remains within $[0, 1]$.

\item The exact-match audit in Table~\ref{tab:exactmatch} does not separate
memorisation from a word-length effect. Short words are both easier to
transcribe correctly and more likely to be produced by the akshara-level
grammar, so the two explanations cannot be told apart from these counts alone.

\end{enumerate}

\section{Conclusion}
\label{sec:conclusion}

In this paper we measured test character error rate as a function of
human-transcribed training words for handwritten Devanagari. The setup was as
follows: four initialisation regimes and nine annotation budgets, using six
random seeds at every point. The resulting curves were then converted into
annotation-equivalent terms.

Random initialisation requires roughly four times more transcription to reach a
moderate error rate compared to supervised synthetic pretraining. Supervised
synthetic pretraining is worth approximately 136 transcribed words before any
real label is seen at all. As the target accuracy improves this advantage gets
smaller, and at the most demanding target we measure it cannot be distinguished
from no saving at all. We separated the pretraining task from how much of the
network was transferred, and observed that most of the benefit at small
annotation budgets comes from the transferred sequence model and output head
rather than from the encoder; however, this sequence-model contribution is short
lived, while the smaller encoder contribution remains detectable over a wider
range of budgets. Masked image modelling showed negative transfer when compared
to random initialisation, but only over a bounded range of budgets, from $n=50$
to $n=500$, and under the compute budget described in
Section~\ref{sec:regimes}.

Thus, the contribution of this study is a fine-grained measurement of what these
methods are worth, with an uncertainty interval at every point, in the units
that matter for planning a transcription project: human annotation eff
\section*{Code Availability}
Code to reproduce the synthetic corpus, the pretraining and fine-tuning sweep,
and every table and figure in this paper is available at
\url{https://github.com/kpandeymabglesh-1224/devanagari-annotation-efficiency}.
The \texttt{results/} directory in that repository is the authoritative source
for every number reported here. The IIIT-HW-Dev dataset is not redistributed;
it is available from CVIT, IIIT Hyderabad under its own terms, and the
repository documents the expected layout.
\begin{ack}
Claude (Anthropic) was used for manuscript editing, code development,
refinement and debugging. All experiments were designed, executed, and verified
by the authors, who take full responsibility for the contents.
\end{ack}

\bibliographystyle{plainnat}
\bibliography{references}

\end{document}